\documentclass[journal]{IEEEtran}
\IEEEoverridecommandlockouts

\usepackage{cite}
\usepackage[hidelinks]{hyperref}

\usepackage{amsmath,amssymb,amsfonts}
\usepackage{algorithmic}
\usepackage{graphicx}
\usepackage{textcomp}
\usepackage{xcolor}
\usepackage{gensymb}
\usepackage{amssymb}
\usepackage{tabularray}
\usepackage{caption}
\usepackage{subcaption}
\newcommand{\simimage}{\mathbf{x}}
\newcommand{\depthimage}{\mathbf{d}}
\newcommand{\labelimage}{\mathbf{c}}
\newcommand{\latent}{\mathbf{z}}
\newcommand{\ytilde}{\mathbf{\tilde{y}}}
\newcommand{\yhat}{\mathbf{\hat{y}}}
\newcommand{\realimage}{\mathbf{y}}
\newcommand{\ytildep}{\mathbf{\tilde{y}_{p}}}
\newcommand{\ytildem}{\mathbf{\tilde{y}_{m}}}
\newcommand{\ytildel}{\mathbf{\tilde{y}_{l}}}
\newcommand{\yprimem}{\mathbf{y^{\prime}_{m}}}

\newcommand{\G}{\mathbf{G}}
\newcommand{\bigy}{\mathbf{Y}}
\newcommand{\bigx}{\mathbf{X}}
\newcommand{\Genc}{\mathbf{G_{enc}}}
\newcommand{\Gdec}{\mathbf{G_{dec}}}
\newcommand{\D}{\mathbf{D}}
\newcommand{\E}{\mathbf{E}}
\newcommand{\Hnet}{\mathbf{H}}
\newcommand{\RS}{\mathbf{RS}}
\title{\LARGE \bf A Unified Generative Framework for Realistic Lidar Simulation in Autonomous Driving Systems}
\author{Hamed Haghighi$^{1}$, Mehrdad Dianati$^{2}$, Valentina Donzella$^{1}$ and Kurt Debattista$^{1}$%
\thanks{This research is supported in part by the University of Warwick's Centre for Doctoral Training in Future Mobility Technologies and in part
by the Hi-Drive Project through the European Union’s Horizon 2020 Research
and Innovation Program under Grant Agreement No 101006664.}
\thanks{$^{1}$H. Haghighi, K. Debattista, and V. Donzella are with WMG, University of Warwick, Coventry, U.K. (Corresponding author: Hamed.Haghighi@warwick.ac.uk)
}
\thanks{$^{2}$M. Dianati is with the School of Electronics, Electrical Engineering and Computer Science at Queen’s University of Belfast and WMG at the University of Warwick.
}
}

\begin{document}

\maketitle
\thispagestyle{empty}
\pagestyle{empty}

\begin{abstract}
Simulation models for perception sensors are integral components of automotive simulators used for the virtual Verification and Validation (V\&V) of Autonomous Driving Systems (ADS). These models also serve as powerful tools for generating synthetic datasets to train deep learning-based perception models. Lidar is a widely used sensor type among the perception sensors for ADS due to its high precision in 3D environment scanning. However, developing realistic Lidar simulation models is a significant technical challenge. In particular, unrealistic models can result in a large gap between the synthesised and real-world point clouds, limiting their effectiveness in ADS applications. Recently, deep generative models have emerged as promising solutions to synthesise realistic sensory data. However, for Lidar simulation, deep generative models have been primarily hybridised with conventional algorithms, leaving unified generative approaches largely unexplored in the literature. Motivated by this research gap, we propose a unified generative framework to enhance Lidar simulation fidelity. Our proposed framework projects Lidar point clouds into depth-reflectance images via a lossless transformation, and employs our novel \textbf{Co}ntrollable \textbf{Li}dar point cloud \textbf{Gen}erative model, CoLiGen, to translate the images.  We extensively evaluate our CoLiGen model, comparing it with the state-of-the-art image-to-image translation models using various metrics to assess realness, faithfulness, and performance of a downstream perception model. Our results show that CoLiGen exhibits superior performance across most metrics. The dataset and source code for this research are available at \url{https://github.com/hamedhaghighi/CoLiGen.git}.
\end{abstract}
\begin{IEEEkeywords}
 deep learning, deep generative models, GANs, realistic Lidar simulation, image-to-image translation, contrastive learning, and autonomous driving systems.
\end{IEEEkeywords}

\section{Introduction} \label{introdcution}
Virtual verification and validation (V\&V) have become increasingly crucial elements of safety assurance frameworks for Autonomous Driving Systems (ADS). These processes significantly reduce the inherent safety risks and prohibitive costs associated with extensive field tests and trials \cite{Kalra2016}. To ensure the efficacy of virtual V\&V, the development of highly realistic simulation models, particularly for perception sensors, is essential.  Simulation models for perception sensors not only play an important role in virtual V\&V but also act as invaluable tools for generating large datasets needed to train deep learning-based perception models \cite{shift}. To this end, this paper focuses on developing a realistic simulation model for Lidar, one of the most widely used perception sensors in ADS due to its ability to capture precise 3D environmental data \cite{s19030648}.
\par
The deployment of commonly used perception sensor simulators and the synthetic datasets they produce poses a significant technical challenge: bridging the complex discrepancy referred to as the `sim-to-real gap' \cite{Kadian2019Sim2RealPD}. This disparity emerges from the difficulties in accurately replicating 3D assets, light propagation, environmental conditions, and sensory effects, leading to noticeable differences between the quality of simulated sensory data and its real-world counterpart. Such flaws can undermine the effectiveness of virtual V\&V, resulting in unreliable safety assurance practices \cite{10.1002/aaai.12141}. Moreover, a wide sim-to-real gap can reduce the usefulness of synthetic sensory data when used to train machine learning models for ADS perception and planning subsystems \cite{Triess2021ASO}.

\par
While high-fidelity simulation models can be developed using complex physics-based approaches, such models require substantial computational resources. This presents a major problem for ADS applications that require high-speed simulation models to accelerate the virtual V\&V processes. Consequently, there has been a growing interest in data-driven methods to bridge the sim-to-real gap \cite{9564034,Triess2021ASO}. Among these, deep generative models have recently shown remarkable capabilities in synthesising realistic high-dimensional data. In Lidar simulation, deep generative models are primarily used in a hybrid manner and integrated within traditional simulation pipelines to model complex Lidar data attributes. For instance, Vacek et al.  \cite{9339933} designed a U-Net \cite{Ronneberger2015} model to learn Lidar reflectance from the real-world KITTI dataset \cite{Geiger2013}, implementing it within the CARLA simulator \cite{Dosovitskiy2017}. Despite promising results, these methods often focus on specific Lidar attributes, relying on conventional simulation techniques for the other aspects. This limitation has led to the emergence of unified methods that translate all data attributes into more realistic representations. However, the unified generative methods for Lidar simulation remain relatively unexplored. The existing approaches use Bird's Eye View (BEV) representations of Lidar point clouds \cite{Saleh2019DomainAF,10.1109/ITSC48978.2021.9564553}, which prevents them from generating the original 3D structure. Furthermore, these models rely on the vanilla CycleGAN's \cite{cyclegan} mapping framework and consistency loss for translation which often leads to unrealistic results, as discussed in Section \ref{subsec:comparison to SOTA}. Lastly, these models are primarily leveraged for improving downstream tasks' performance\cite{xiao2022transfer}, without comprehensively investigating different aspects of simulation fidelity.

\par
\begin{figure*}[t!]
\centering
\includegraphics[width=\textwidth]{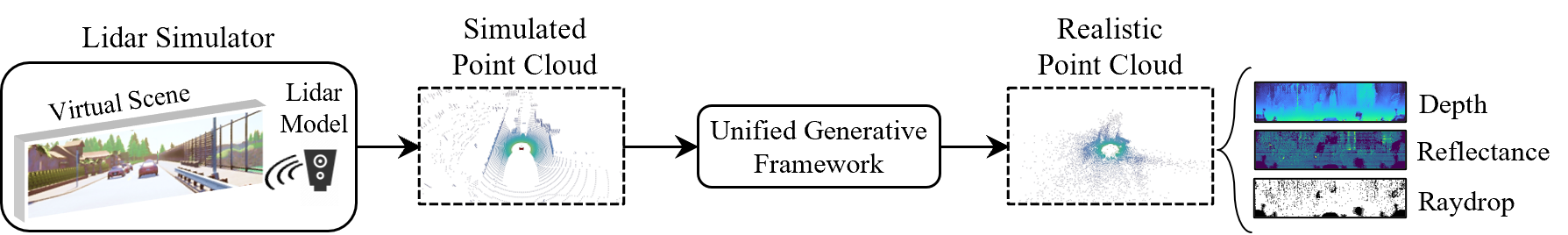}
\caption{We introduce a unified generative framework to enhance the realism of Lidar simulation. Our framework directly translates all Lidar data attributes into more realistic representations, enhancing both the shape characteristics, derived from the points' depth and raydrop pattern, as well as reflectance properties. This approach substantially improves the overall fidelity of Lidar simulation.}
\label{fig:teaser}
\end{figure*}

In this paper, we introduce a novel unified generative framework designed to improve the realism of Lidar simulation, as illustrated in Fig. \ref{fig:teaser}. Our framework uses a lossless transformation to project Lidar point clouds into depth-reflectance images, hence enabling the reconstruction of the original data structure. Subsequently, the images are translated into more realistic representations using our proposed \textbf{Co}ntrollable \textbf{Li}dar point cloud \textbf{Gen}erative model, referred to as CoLiGen. CoLiGen is the first model of its kind capable of translating all essential Lidar data attributes, including shape features derived from points' depth and raydrop pattern, as well as reflectance properties. Furthermore, our model undergoes a comprehensive evaluation that extends the current scope of research in the field, utilising a variety of metrics to assess simulation fidelity. Within our CoLiGen model, we adopt and demonstrate the effectiveness of two techniques: the decomposition of raydrop synthesis, inspired by its success in the unconditional generation task \cite{Nakashima2021LearningTD}, and contrastive learning, motivated by its success in enforcing consistency for unpaired image translation \cite{park2020cut}.  
\par
To evaluate the performance of the proposed CoLiGen, we created a synthetic dataset named semantic-CARLA, which consists of point-wise annotated Lidar point clouds and RGB camera images generated using the CARLA simulator \cite{Dosovitskiy2017}. An extensive evaluation of CoLiGen is conducted in this paper from the perspectives of realness, faithfulness, and a downstream perception model. The results demonstrate that CoLiGen surpasses state-of-the-art (SOTA) models, as measured by key qualitative and quantitative performance indicators such as Fréchet Inception Distance (FID) \cite{Heusel2017}. In summary, the main contributions of this paper include the following.
\begin{itemize}
\item We introduce CoLiGen, a novel controllable generative model designed to translate all Lidar data attributes for enhanced realism. CoLiGen outperforms SOTA image-to-image translation models across most evaluation metrics.
\item We conduct a comprehensive evaluation of the CoLiGen model in terms of realness, faithfulness, and the performance of a downstream perception model.
\item We highlight the importance of decomposing raydrop synthesis and utilising contrastive learning in improving the quality of the synthesised point clouds, with the former aiding in capturing sparsity patterns and the latter ensuring consistency across unpaired translation.
\item We introduce semantic-CARLA, a new publicly accessible dataset consisting of synthetic, point-wise annotated Lidar point clouds and RGB images of driving scenes.
\end{itemize}

The remainder of the paper is organised as follows: Section \ref{related-work} reviews relevant literature; Section \ref{sec:problem_formulation} formulates the problem; Section \ref{sec:gen-framework} details the proposed framework; Section \ref{sec:evaluations} presents a comprehensive performance evaluation of the proposed framework, and Section \ref{sec:conclusion} concludes the paper and discusses the potential future research directions.

\section{Related Work} \label{related-work}
This section presents an overview of deep generative models for Lidar point cloud synthesis and synthetic Lidar datasets found in the literature.
\subsection{Deep Generative Models for Lidar Point Cloud Synthesis} \label{subsec:gen-lidar-models}
Generative models, particularly those leveraging deep learning techniques, have fundamentally transformed the approach to modelling high-dimensional data, outperforming traditional distribution matching methods. While these models have primarily been employed for the synthesis of photo-realistic RGB images, their application has expanded to include the generation of Lidar point clouds. The literature outlines two distinct categories of Lidar point cloud generative models: unconditional and controllable. The subsequent sections discuss these categories in detail.
\subsubsection{Unconditinal Synthesis}
Unconditional models aim to capture the inherent distribution of Lidar point clouds without relying on external conditions or guidance. While these models cannot be directly applied to ADS, they form the basis for more sophisticated models. The majority of unconditional models have adapted widely used image generative models, including Generative Adversarial Networks (GANs) \cite{10.5555/2969033.2969125}, denoising diffusion models \cite{10.5555/3495724.3496298}, and auto-regressive transformers \cite{10.5555/3295222.3295349}. The following paragraphs separately elaborate on these models.
\par
Several pioneering works have adapted GANs to capture the distribution of Lidar point clouds. Caccia et al. \cite{Caccia2018DeepGM} unravelled Lidar point clouds into range images and adapted the Deep Convolutional GAN (DCGAN) \cite{Radford2015UnsupervisedRL} architecture for synthesis. Subsequent frameworks, such as DUSty \cite{Nakashima2021LearningTD} and DUSty2 \cite{nakashima2022generative}, incorporated ray-drop estimation and Implicit Neural Representation (INR) into the GAN training pipeline, achieving more realistic results and expanding the model's applicability to data restoration and up-sampling tasks.
\par
Recent advancements in denoising diffusion models \cite{10.5555/3495724.3496298} have led to their successful application in unconditional Lidar point cloud generation, with notable contributions from LiDARGen \cite{10.1007/978-3-031-20050-2_2}, R2DM \cite{nakashima2024lidar}, RangeLDM \cite{hu2024rangeldm}, and LiDM \cite{ran2024towards}. These models introduce various techniques to ensure physical realism, stable training, high-quality outputs, and better alignment with real-world data. Furthermore, auto-regressive transformers \cite{10.5555/3295222.3295349} and VQ-VAE \cite{NIPS2017_7a98af17} have been incorporated into models such as UltraLiDAR \cite{xiong2023learning} and LidarGRIT \cite{haghighi2024taming}, enabling step-by-step generation and precise noise modelling through repeated sampling in the latent space. Despite the exceptional performance of these unconditional models in generating realistic Lidar point clouds, their inability to control the synthesis of specific driving scenarios highlights the necessity for further research into controllable generation methods.

\subsubsection{Controllable Synthesis}
 In contrast to unconditional models that generate data stochastically, controllable models utilise auxiliary inputs to guide the generation process, enabling their integration into automotive simulation frameworks. In the area of Lidar simulation for ADS, controllable generative models can be divided into two categories: hybrid and unified methods. Hybrid methods integrate generative components within the Lidar simulation pipeline to simulate specific attributes, such as Lidar reflectance. Conversely, unified approaches use a cohesive framework to simultaneously generate all data attributes, translating both 3D shape and ray reflectance properties.
 \par
Hybrid methods often use deep generative models to synthesise data attributes that are computationally complex to simulate realistically, \textit{e.g.} Lidar reflectance. For instance, Vacek et al. \cite{9339933} and Wu et al. \cite{Wu2018} used neural networks to predict Lidar reflectance based on spatial coordinates. Similarly, Guillard et al. \cite{guillard2022learning}, Zhao et al. \cite{Zhao2020ePointDAAE}, and Manivasagam et al. \cite{Manivasagam2020} focused on estimating raydrop, i.e. dropout/noise. Although these methods yield promising results, they are constrained by their focus on specific attributes and still depend on traditional simulation algorithms for other aspects. \par

Conversely, unified approaches aim to synthesise all data attributes by uncovering a mapping between the simulated and real domains. Since real and simulated data are typically unpaired during the training of these models, the problem can be framed as finding an unsupervised domain mapping. Sallab et al. \cite{Sallab2019} and Barrera et al. \cite{10.1109/ITSC48978.2021.9564553} converted Lidar point clouds into BEV range images and adapted the CycleGAN \cite{cyclegan} to map the images. Both studies demonstrated a superior performance on the object detection task while using the translated BEV images. More recently, Xiao et al. \cite{xiao2022transfer} tackled sim-to-real domain gaps in semantic segmentation by proposing a Point Cloud Translation (PCT) model working with the 3D representation of point clouds.

Our CoLiGen model shares the goal of unified approaches in encompassing all data attributes in translation. However, it distinguishes itself by employing a lossless representation of Lidar point clouds based on polar coordinates, in contrast to the non-bijective BEV transformation. Additionally, it incorporates innovative techniques such as decomposed raydrop synthesis \cite{nakashima2021learning} and contrastive consistency loss \cite{park2020cut}, rather than relying solely on the vanilla CycleGAN framework \cite{cyclegan}. These enhancements result in superior performance compared to SOTA image-to-image translation models, such as GcGAN~\cite{8953926}.

\subsection{Lidar Datasets} \label{subsec:lidar-datasets}
Collecting real-world Lidar data has recently gained significant momentum due to the success of data-driven driving functions. Several datasets such as Semantic-KITTI \cite{behley2019iccv}, semantic-POSS \cite{pan2020semanticposs}, Nuscenes \cite{nuscenes}, and Waymo \cite{Sun_2020_CVPR}, have been provided, containing a diverse set of deriving scenes and conditions. The datasets have been used as benchmarks for different perception tasks such as object detection and semantic segmentation. Despite their large contribution to autonomous driving research, the costs of manual annotation of Lidar point clouds coupled with the risks of collecting edge-case scenarios have motivated the researcher to gather data from simulation environments \cite{song2023synthetic}. \par
Several synthetic driving datasets tailored for perception tasks have been released recently. However, only a few of these contain Lidar data consisting of point-wise semantic labels, RGB colour, and reflectance. Among these, Presil \cite{10.1109/IVS.2019.8813809} stands as one of the pioneering datasets using the GTA-V graphics engine to ray-cast Lidar point clouds. While the dataset contains RGB images and point-wise semantic labels, the classes are only limited to `Car' and `Person' due to the constraints of the GTA-V simulation environment. On the other hand, the more recent datasets such as SHIFT \cite{shift} and AIOD \cite{Weng2020_AIODrive} include comprehensive sensor suits including different levels of annotations. While the SHIFT \cite{shift} dataset focuses mainly on capturing a multitude of discrete and continuous domains in real-world driving, AIOD  covers out-of-distribution scenarios and long-range perception. 
Although the aforementioned datasets mainly focus on providing larger datasets containing more driving conditions, our proposed dataset is mainly created for realistic Lidar simulation. Specifically, we designed our dataset, namely semantic-CARLA, to mirror the scene arrangement, sensor configuration, and data structure of the widely adopted semantic-KITTI \cite{behley2019iccv} dataset. This is to reduce the sim-to-real domain shift, enabling the generative model to focus on the differentiation in object geometries and the simulation physics. Furthermore, this design concept can apply to other datasets, as they all derive from the KITTI dataset and adhere to a similar structure.

\begin{figure*}[t!]
\centering
\includegraphics[width=\textwidth]{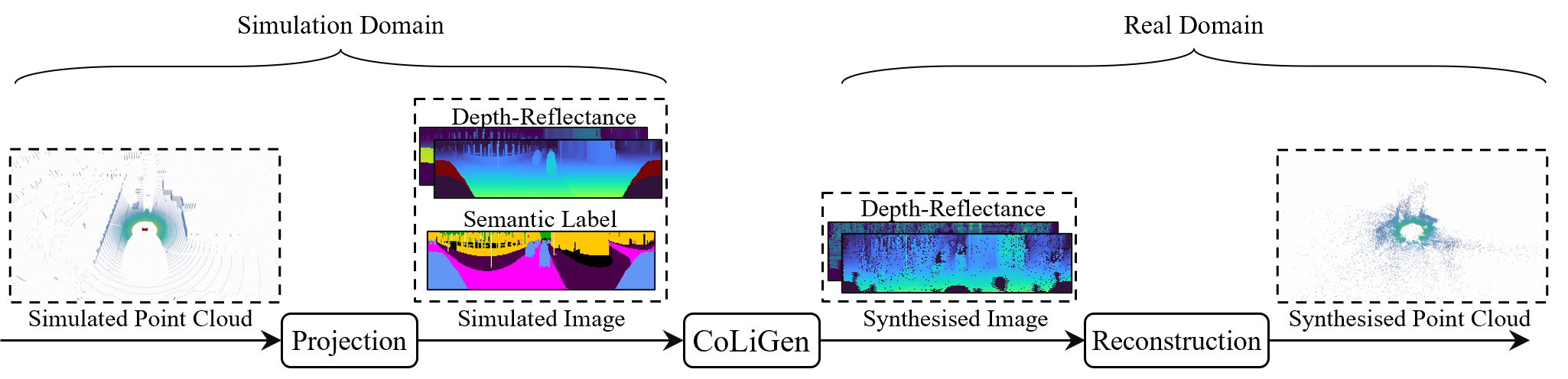}
\caption{Overview of the proposed generative framework. Our framework begins by projecting the simulated point cloud into depth-reflectance images and a semantic label layout. Subsequently, these representations are translated using our novel CoLiGen model to enhance their realism. Finally, the more realistic point cloud is reconstructed from the synthesised depth-reflectance images, thus completing the process.}
\label{fig:entire_framework}
\end{figure*}

\section{Problem Formulation} \label{sec:problem_formulation}
Our goal is to learn a domain mapping function, denoted by $G: \mathcal{P_{X}} \rightarrow \mathcal{P_{Y}}$, that translates Lidar point clouds from simulation domain $\mathcal{P_{X}} \subset \mathbb{R}^{N \times (4+M)}$ into real domain $\mathcal{P_{Y}} \subset \mathbb{R}^{N\times 4}$. This learning process is performed using the corresponding simulated $P_{X} = \{p_{\simimage} \in \mathcal{P_{X}}\}$ and real datasets $P_{Y} = \{p_{\realimage} \in \mathcal{P_{Y}}\}$. In this formulation, simulated Lidar point cloud $p_{\simimage}$ consists of $N$ points with four plus $M$ dimensions. The initial four dimensions include three dimensions for 3D spatial coordinates and one for reflectance; While additional $M$ dimensions contain auxiliary data such as point-wise semantic labels or RGB colours. Real Lidar point cloud $p_{\realimage}$ contains N points with four dimensions (for 3D coordinates and reflectance). As detailed in Section \ref{subsec:data projection}, we project the Lidar point clouds into images of dimensions $H \times W$. Using this data representation, the problem formulation is transformed into an equivalent of finding a mapping $G$ between the projected Lidar point clouds from simulation $\mathcal{X} \subset \mathbb{R}^{H \times W \times (2+M)}$ and real domain $\mathcal{Y} \subset \mathbb{R}^{H \times W \times 2}$. This is given the corresponding simulated $\bigx = \{\simimage \in 
\mathcal{X}\}$ and real dataset $\bigy = \{\realimage \in \mathcal{Y}\}$.
\par

\section{Proposed Generative Framework} \label{sec:gen-framework}
The overview of the proposed generative framework is depicted in Fig. \ref{fig:entire_framework}. We first project the simulated Lidar point clouds, into depth-reflectance images and semantic label images. These images are then translated into highly realistic depth-reflectance images through our CoLiGen model. Finally, the output of the model is converted back into realistic Lidar point clouds through a reconstruction transformation. The details of the components, as well as the training process of the CoLiGen framework are discussed in the following subsections.


\subsection{Point Cloud Projection and Reconstruction} \label{subsec:data projection}
To leverage the power of CNN-based architectures in our CoLiGen model, we convert the Lidar point clouds into image-based representation via the spherical transformation. Considering $H, W$ as the vertical and horizontal angular resolution of mechanical rotating Lidar, we create an image of $H \times W$ dimension, assigning each point a pixel based on its azimuth and elevation angle $(\theta,\phi) \in \mathbb{R}^{2}$. Point-wise depth, reflectance, and M auxiliary attributes are stored in the assigned pixel. Therefore, the original point cloud array with a shape of $N \times (4+M)$ is transformed into an image array with a shape of $H \times W \times (2+M)$ using this projection. The angles $(\theta, \phi)$ can be calculated using 3D spatial coordinates $(x,y,z) \in \mathbb{R}^{3}$ of point cloud as:
\begin{equation}
(\theta, \phi) = (\arctan2(y, x) , \arctan2(z, \sqrt{x^2 + y^2})).
\end{equation}
Similarly, we reconstruct the output depth-reflectance image 
 array with a shape of $H \times W \times 2$ back to the point cloud array with a shape of $N \times 4$ in a lossless manner. The reconstruction utilises the azimuth and elevation angle $(\theta, \phi)$ of each pixel along with the corresponding depth $d \in \mathbb{R}$, to restore the $(x,y,z)$ as:
\begin{equation}
(x,y,z)=(d\cos(\phi)\cos(\theta),d\cos(\phi) \sin(\theta), d\sin(\phi)).
\end{equation}

\subsection{CoLiGen Model} \label{sbusec:network}
\begin{figure*}[t!]
\centering
\includegraphics[width=\textwidth]{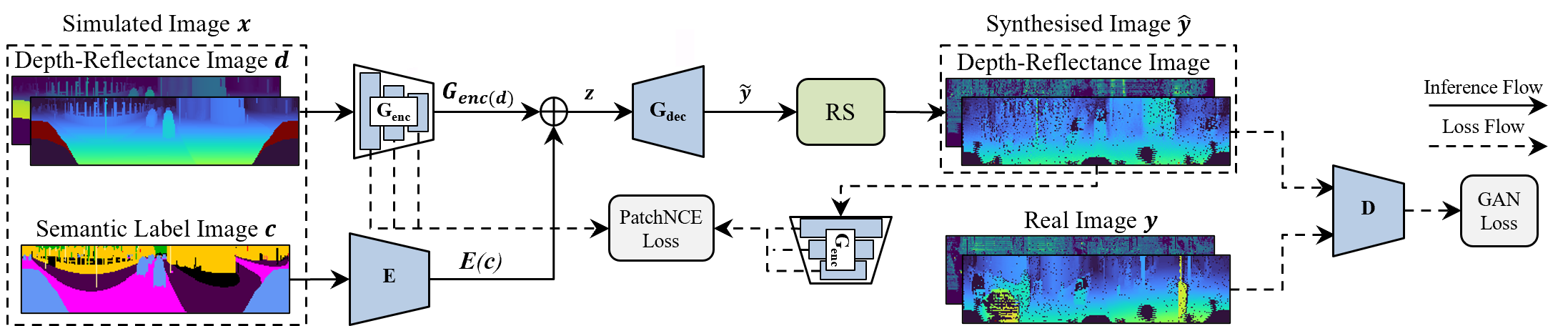}
\caption{Overview of the inference and training losses of our CoLiGen model. The model consists of $\Genc$, $\E$, and $\Gdec$ sub-networks. $\Genc$ encodes depth-reflectance image $\depthimage$ and $\E$ encodes auxiliary image $\labelimage$. These two codes are summed up to form $\latent$ which are then decoded by $\Gdec$ to output $\ytilde$. The Raydrop Synthesis (RS) module renders raydrop and synthesises the final image $\yhat$. The network is trained with GAN loss \cite{10.5555/2969033.2969125} using the discriminator $\D$, and PatchNCE loss \cite{park2020cut} using the features extracted by $\Genc$.}
\label{fig:training_diag}
\end{figure*}

Inspired by the recently developed image-to-image translation models \cite{alotaibi2020deep}, our CoLiGen model $\G$ adopts an encoder-decoder architecture, as shown in Fig. \ref{fig:training_diag} (Inference Flow). The inference flow of $\G$ entails encoding the simulated images $\simimage$ with $\Genc$ and $\E$, decoding the latent vector with $\Gdec$, and then using Raydrop Synthesis (RS) to synthesise the realistic output $\yhat$. Since we needed to use $\Genc$ to extract features from both $\simimage$ and $\yhat$ for the PatchNCE loss (see Section \ref{subsec:training} for more details), and these inputs may not necessarily have similar tensor shapes, we decouple the encoding of $\simimage \in \mathbb{R}^{H \times W \times 3}$ into the encoding of depth-reflectance image $\depthimage \in \mathbb{R}^{H \times W \times 2}$ and the encoding of the semantic label image $\labelimage \in \mathbb{R}^{H \times W \times 1}$.

In this design, $\Genc$ and $\E$ encodes $\depthimage$ and $\labelimage$ respectively and form latent vector $\latent = \Genc(\depthimage) + \E(\labelimage)$. The vector $\latent$ is decoded to $\ytilde$ using $\Gdec$, and finally,  $\ytilde$ is converted to $\yhat$ using the RS component. The mapping process can be described as:
\begin{equation}
\begin{split}
    \yhat&=\G(\simimage) = \RS(\ytilde)= \RS(\Gdec(\latent))\\
     &= \RS(\Gdec(\Genc(\depthimage) + \E(\labelimage))) .
\end{split}
\end{equation}
\par
\begin{figure}[t!]
\centering
\includegraphics[width=\linewidth]{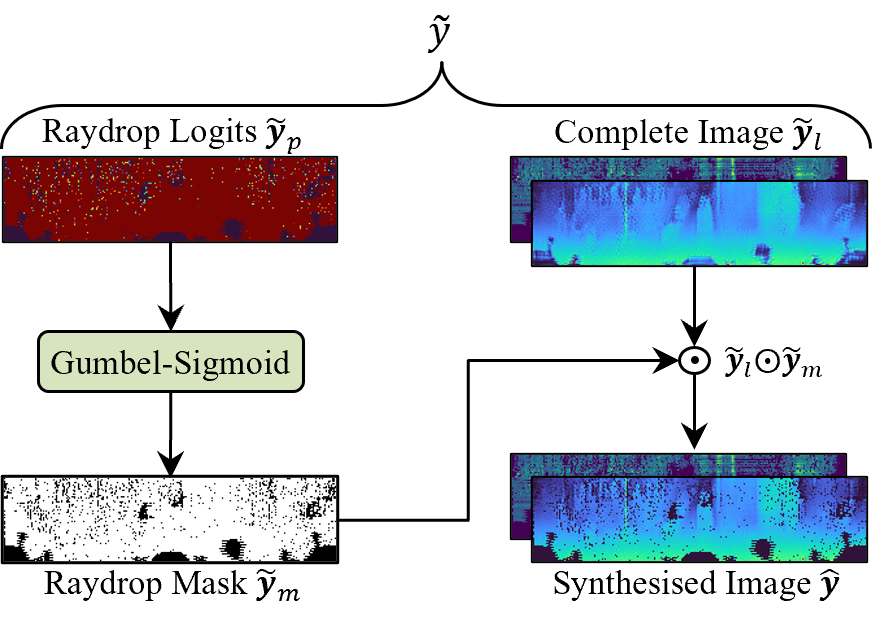}
\caption{Overview of Raydrop Synthesis (RS) module. The RS module inputs $\ytilde$ consisting of raydrop logits $\ytildep$ and complete depth-reflectance image $\ytildel$. The Gumbel-Sigmoid function \cite{journals/corr/JangGP16} is applied to raydrop logits, yielding the binary raydrop mask $\ytildem$. Lastly, the raydrop mask is multiplied with the complete image, resulting in the final synthesised image~$\yhat$. }
\label{fig:RS}
\end{figure}

As shown in Fig. \ref{fig:RS}, the $\RS$ component renders raydrop patterns, i.e. missing transmitted rays, on the synthesised depth-reflectance image $\ytilde$, Raydrop happens when the ray travel distance surpasses the maximum detection range or when it hits refractive or highly reflective surfaces. Concerning the image-based representation of the Lidar point cloud, raydrop can be represented as a binary mask. While ray-drop mask can be implicitly synthesised by our generator, CNNs commonly encounter difficulty in modelling stochastic noise in images \cite{Bora2018AmbientGANGM}. Therefore, inspired by the successful decomposed synthesis of raydrop in unconditional generation \cite{Nakashima2021LearningTD}, we disentangle the synthesis of Lidar scans in our controllable setting by modelling complete scans (without the missed information) and the ray-drop mask independently. Via this technique, image $\ytilde \in \mathbb{R}^{H \times W \times 3}$ is first generated which is a concatenation of complete depth-reflectance images $\ytildel \in \mathbb{R}^{H \times W \times 2}$ and raydrop logits image $\ytildep \in \mathbb{R}^{H \times W \times 1}$. Then the ray-drop mask $\ytildem$ is obtained by sampling from a Bernoulli distribution with $\ytildep$ as the parameter:
\begin{equation}
\begin{split}
    \ytildem \sim Bernoulli(\ytildep),
\end{split}
\end{equation}
where $\ytildem$ is a binary mask indicating which points to keep and which to drop. Using the mask, $\yhat$ is calculated by multiplying the mask into the complete scans as $\yhat = \ytildem \odot \ytildel$.  
Since sampling from the Bernoulli distribution is not a differentiable operation, the re-parametrisation trick is used to estimate the gradient in the backpropagation. $\ytildem$ is re-parameterised with a continuous relaxation $\yprimem$ using Gumbel-Sigmoid distribution \cite{journals/corr/JangGP16} as:
\begin{equation}
\begin{split}
    \yprimem = sigmoid\bigg( \frac{(\log(\ytildep) + \mathbf{g_{1}}) - (\log(1 - \ytildep) + \mathbf{g_{2}})}{\alpha}\bigg),
\end{split}
\label{eq: gumble-sigmoid}
\end{equation}

\noindent where $\mathbf{g_{1}} \in R^{H \times W}$ and $\mathbf{g_{2}} \in R^{H \times W}$ are two i.d.d samples for from $Gumbel(0, I)$ making the process random. Furthermore, $\alpha \in \mathbb{R}$ controls the gap between the relaxed and original Bernoulli distribution. The re-parameterised sample $\yprimem$ is then discretised to binary mask by being thresholded against $0.5$. This can be written as:

\begin{equation}
	\ytildem=\begin{cases}
	1 & \yprimem\geq0.5 \\
	0 & \yprimem<0.5 .
	\end{cases}
\end{equation}
The thresholding is only used for the forward pass of the network and is replaced with the identity function in order to enable gradient propagation. 

\subsection{Training} \label{subsec:training}
\begin{figure}[t!]
\centering
\includegraphics[width=\linewidth]{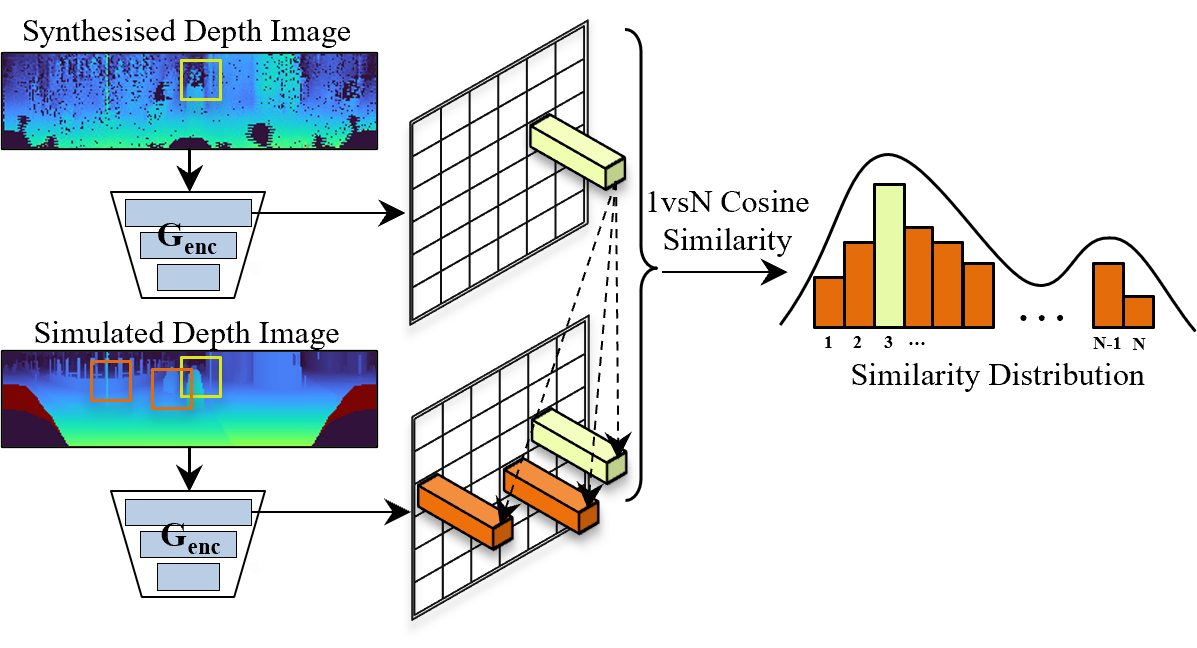}
\caption{Overview of the patchNCE\cite{park2020cut} loss.  We adopt a patch-wise contrastive objective to enforce consistency during the unpaired image translation. This objective ensures that a patch in the image synthesised by the generator's $\G$ has the highest similarity to the patch of the simulated image in the same position, compared to patches in other positions.}
\label{fig:CL}
\end{figure}

As shown in Fig. \ref{fig:training_diag} (Loss Flow), we train our model G using GAN \cite{10.5555/2969033.2969125} and PatchNCE\cite{park2020cut} losses. To synthesise realistic data, the output of the model $\G$ should be indistinguishable from that of real domain data $y$. To this end, we leverage the GAN loss, involving the simultaneous optimisation of two networks, the generator $\G$ and the discriminator $\D$, in a competitive setting. 
The network $\G$ attempts to generate realistic data which is indistinguishable by $\D$, and $\D$ aims to differentiate $\G$'s output from real inputs.
The GAN loss can be calculated as:
\begin{equation}
\begin{split}
    \mathcal{L}_{GAN}(\G, \D, \bigx, \bigy) = &\mathbb{E}_{\realimage \sim \bigy}[\log \D(\realimage)] \\+
                 &\mathbb{E}_{\simimage \sim \bigx}[\log (1-\D( \G(\simimage))]\label{cGAN_equation}.\\
\end{split}
\end{equation}
If the training of $\G$ relies solely on minimising the GAN loss, the model might learn to generate data that is realistic enough to fool the discriminator irrespective of the input. To avoid this problem and to assure that the content of the input is accurately translated,  CoLiGen uses PatchNCE loss \cite{park2020cut}. PatchNCE enforces consistency between input and output images using contrastive learning. Via this technique, the generator learns to synthesise images such that input-output patches at the same spatial location are given the highest similarity compared with patches at the other locations (see Fig. \ref{fig:CL}). For example, an output image patch containing a car matches more closely with the input patch containing a car than with a patch containing a pedestrian. To extract patch-wise information, we use the features already extracted in the encoding layers of the $\G$, denoted as $\Genc$, and post-process it via a Multi-Layer-Perceptron (MLP) $\Hnet$. For an $L$ layer $\Genc$, the features in the layer $\l$ can be described as $\{\latent_{l}\}_{l=1}^{L} = \{ \Hnet_{l}(\mathbf{G^{l}_{enc}}(\simimage))\}_{l=1}^{L}$ for the input images  and $\{\latent_{l} \}_{l=1}^{L} = \{\Hnet_{l}(\mathbf{G^{l}_{enc}}(\yhat)\}_{l=1}^{L}$ for the output images. Each spatial location $u \in \{ 1, \  ...,\ U_{l} \}$ of $\latent_{l}$ contains features of an input patch and the higher the encoding layer $l$ is, the larger patch it captures. We refer to $u^{th}$ location of $\latent_{l}$ as $\latent_{l}^{u} \in \mathbb{R}^{C_{l}}$ and other locations as $\latent_{l}^{U\backslash u} \in \mathbb{R}^{C_{l}}$, where $C_{l}$ is the number of channels at each layer. Cosine similarity is used to establish the correspondence between $\hat{\latent_{l}^{u}}$ and $\latent_{l}^{u}$ as:
\begin{equation}
\begin{split}
    &l(\hat{\latent}_{l}^{u}, \latent_{l}^{u},\latent_{l}^{U\backslash u})=\\ -&\log{\Bigg[\frac{\exp{(\hat{\latent}_{l}^{u} \cdot \latent_{l}^{u} / \tau)}}{\exp{(\hat{\latent}_{l}^{u} \cdot \latent_{l}^{u} / \tau)} + \sum_{s \in U\backslash u} \exp(\hat{\latent}_{l}^{u} \cdot \latent_{l}^{s}/ \tau)} \Bigg]},
\end{split}
\label{eq: CE}
\end{equation}
where $\tau=0.07$ is the temperature to scale the distances.  Using the Equation \ref{eq: CE}, the PatchNCE loss can be written as:
\begin{equation}
\mathcal{L}_{patchNCE}
(\G, \Hnet, \bigx)=\mathbb{E}_{\simimage \sim \bigx}\sum_{l=1}^L \sum_{u=1}^{U_l} (\hat{\latent}_{l}^{u}, \latent_{l}^{u},\latent_{l}^{U\backslash u}).
\end{equation}
We also include PatchNCE loss on the real domain $\mathcal{Y}$, referred to as $\mathcal{L}_{patchNCE}(\G, \Hnet, \bigy)$, to ensure that the network does not change the realistic input images. The overall objective $\mathcal{L}$ can be calculated as:

\begin{equation}
\begin{split}
    \mathcal{L} = \mathcal{L}_{GAN}(\G,\D, \bigx, \bigy) &+  \lambda_{nce}\mathcal{L}_{PatchNCE}(\G, \Hnet, \bigx) \\&+  \lambda_{idt}\mathcal{L}_{PatchNCE}(\G, \Hnet, \bigy).
\end{split}
\label{eq: total_loss}
\end{equation}
The hyperparameters $\lambda_{idt}$ and $\lambda_{nce}$ balance the emphasis between the reconstruction error (faithfulness) in the translation process and the proximity to the real distribution (realness) in the GAN loss.

\section{Evaluations} \label{sec:evaluations}
This section describes the evaluations of our proposed CoLiGen model. First, we introduce the experimental settings, followed by a detailed explanation of the evaluations, which includes comparison to the SOTA models, analysis from the perspective of a downstream perception model, and ablation study.
\subsection{Experimental Settings} \label{subsec:experimental settings}
This subsection describes the datasets, including our semantic-CARLA dataset, evaluation metrics used in our study as well as implementation details of the experiments.

\subsubsection{Datasets} \label{subsec:Semantic-carla}
\begin{figure}[t]
\centering
\includegraphics[width=\linewidth]{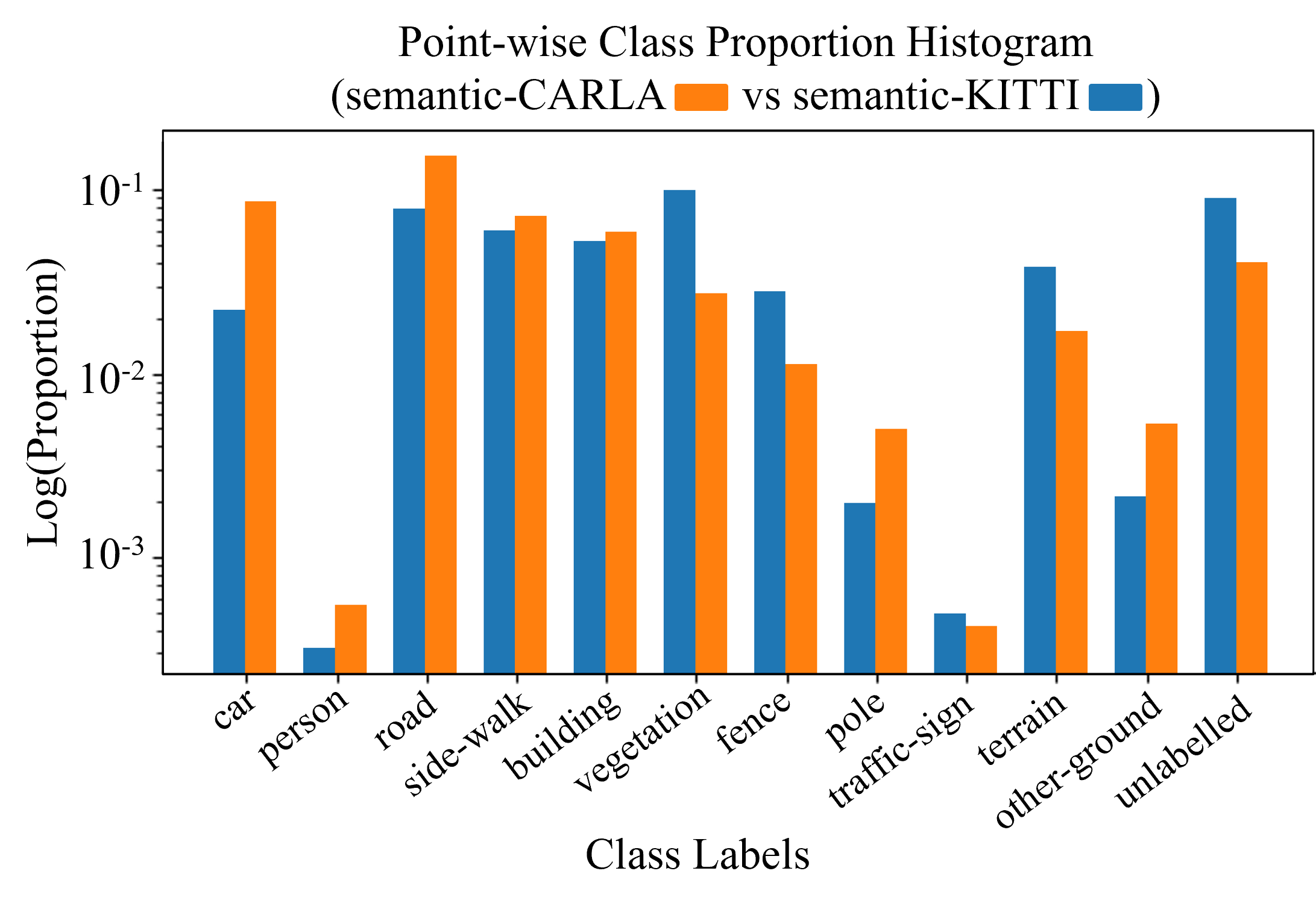}
\caption{Comparison of point-wise class proportion histograms between the semantic-KITTI \cite{behley2019iccv} and our semantic-CARLA datasets. We created the semantic-CARLA dataset to closely align with the semantic-KITTI dataset, ensuring a match in both object classes and their proportions.}
\label{fig:class-proportions}
\end{figure}

The training of CoLiGen necessitates datasets from both simulated and real-world domains to learn the underlying mapping function. For the simulated Lidar point clouds, we developed our own dataset, denoted as semantic-CARLA. We specifically structured this dataset to align with the semantic-KITTI \cite{behley2019iccv} dataset in terms of object classes, point-wise class porpotions (see Fig. \ref{fig:class-proportions}), sensor configuration, and data format. This deliberate design approach aims to minimise the disparity between simulated and real environments, thereby enabling CoLiGen to focus exclusively on identifying sensor-related phenomena. Consequently, our semantic-CARLA dataset is particularly well-suited for research in the field of unsupervised domain mapping for Lidar point clouds. We chose the semantic-KITTI dataset as our real-world reference due to its widespread adoption in autonomous driving research, which has incited the development of subsequent driving datasets.
\par
 \begin{table*}[t]
\centering
\caption{Quantitative comparison to SOTA methods. The best results are bolded, and the second-best results are underlined. The lower the value in all metrics, the better the performance.}
\label{tab:sota}
\begin{tblr}{
  width = \textwidth,
  colsep = 3mm,
  column{3-8}={c},
  cell{2}{1} = {r=6}{},
  cell{8}{1} = {r=6}{}, 
  hline{1-2} = {-}{},
  hline{7,13} = {2-8}{dashed},
  hline{8,14} = {1-8}{},
  hline{1,14} = {-}{1pt}
}
Dataset        & Method                           & FID & SWD & JSD & MMD $\times 10$ & CD $\times 10$ & RMSE\\
Semantic-KITTI & {Semantic-CARLA} & 2980             & 2.35            & 0.25            & 0.14             & -               & -                     \\
               & RINet \cite{guillard2022learning}                           & 2601             & 1.11            & 0.16            & \underline{0.04}     & \underline{0.037}   & 0.380      \\
               & CycleGAN\cite{cyclegan}                         & \underline{2052}     & 0.54            & \underline{0.10}     & 0.08             & \underline{0.037}   & \underline{0.378}                \\
               & GcGAN \cite{8953926}                            & 2458             & \underline{0.46}    & 0.47            & 0.73             & 0.225           & 0.409                   \\
               & CoLiGen (\textbf{Ours})                    & \textbf{1851}    & \textbf{0.45}   & \textbf{0.07}   & \textbf{0.03}    & \textbf{0.032}  & \textbf{0.376}         \\
               & {Training Set}  & 1318             & 0.31            & 0.01            & 0.02             & -               & -                             \\
SemanticPOSS   & {Semantic-CARLA} & 4819             & 2.09             & 0.51             & 0.24             & -               & -                                \\
               & RINet\cite{guillard2022learning}                              & 4019             & 1.33             & \underline{0.27}     & \textbf{0.06}    & 0.070           & \underline{0.258}         \\
               & CycleGAN\cite{cyclegan}                          & 3775             & \underline{1.06}     & 0.36             & 0.45             & \underline{0.066}   & 0.502               \\
               & GcGAN\cite{8953926}                          & \underline{3522}     & 1.52             & 0.41             & 0.38             & 0.114           & 0.347                 \\
               & CoLiGen (\textbf{Ours})                    & \textbf{3365}    & \textbf{0.59}    & \textbf{0.23}    & \underline{0.09}     & \textbf{0.063}  & \textbf{0.217}        \\
               & {Training Set}   & 2848             & 0.75            & 0.06             & 0.04             & -               & -        \\
\end{tblr}
\end{table*}
To create semantic-CARLA, we used the CARLA \cite{Dosovitskiy2017} simulator and replicated the configuration of Lidar and RGB camera sensors to resemble the original setup of the KITTI \cite{Geiger2013} dataset. Specifically, we simulated a 10 frame-per-second (fps) Velodyne HDL-64E Lidar producing approximately 100k points in each rotation, along with a 10 fps RGB camera with $1241 \times 376$ image resolution. We also recorded point-wise semantic labels of Lidar point clouds classifying points into sixteen different classes. The entire dataset consists of approximately 34k scans collected in 8 different CARLA environments.

 Concerning the real datasets, we picked the commonly used Lidar datasets of Semantic-KITTI \cite{behley2019iccv} and semantic-POSS \cite{pan2020semanticposs}.  Semantic-KITTI is a point-wised labelled version of the original KITTI's visual odometry benchmark. The dataset consists of 22 sequences of Lidar (Velodyne HDL-64E) scans, from which 11 sequences (approximately 19k scans) have publicly available annotations. Semantic-POSS have the same data format as semantic-KITTI and contains 3k complex Lidar scans collected at Peking University.  We designated sequences 7 and 8 from the semantic-KITT, as well as sequences 4 and 5 from SemanticPOSS for validation and testing, respectively. 
\begin{figure*}[t]
\centering
\includegraphics[width=1.02\textwidth]{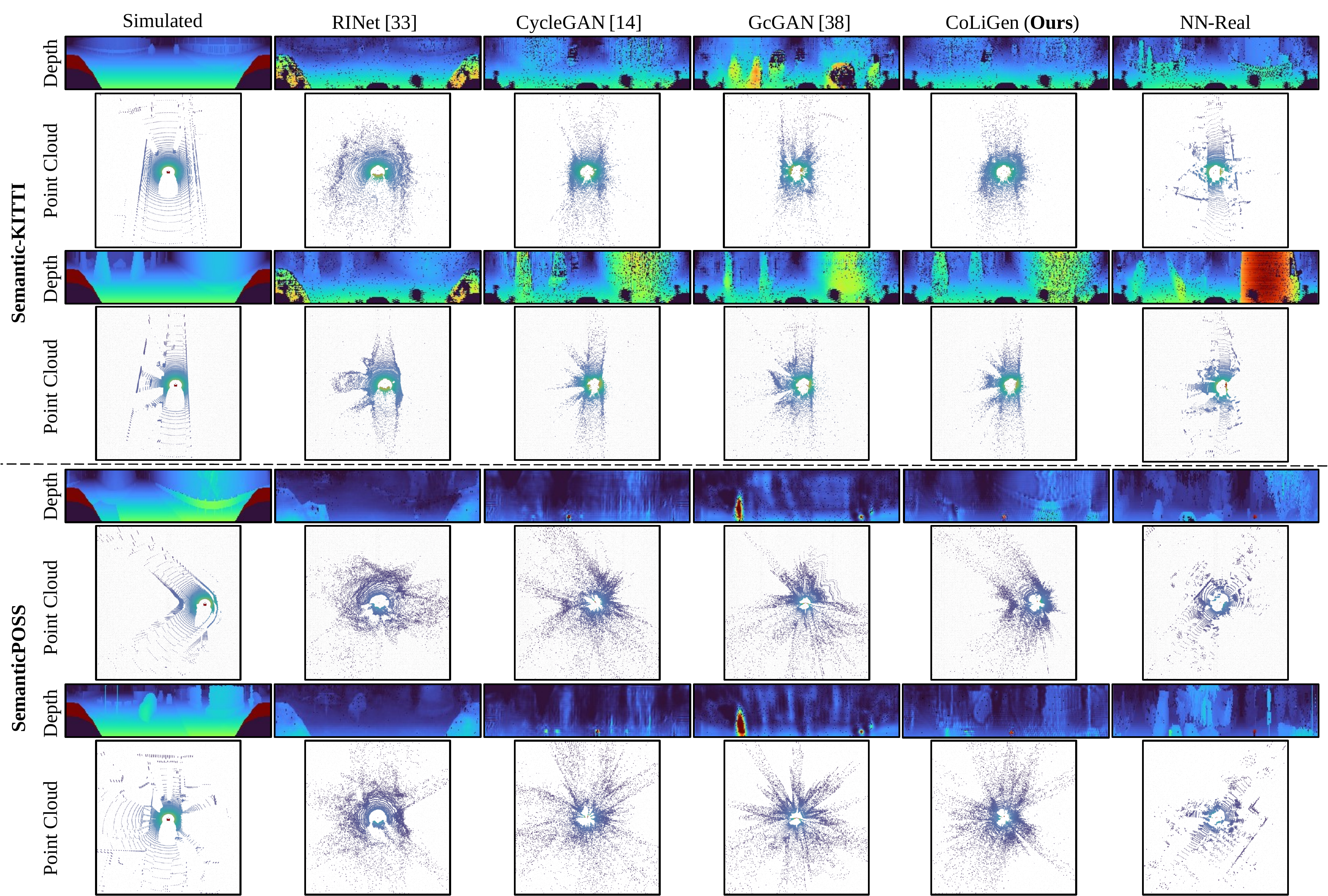}
\caption{Qualitative comparison to SOTA models. Each column demonstrates a different model, and each row shows translated depth images and the reconstructed point clouds in the specified dataset. `Simulated' and `NN-Real' columns exhibit the simulated input data and nearest neighbour sample in the real dataset, respectively. Notably, our CoLiGen model can synthesise more realistic outputs without hallucination (visible in the depth images), while remaining more faithful to the original simulated input (visible in point clouds).}
\label{fig:qual_comp}
\end{figure*}

\subsubsection{Metrics} \label{sec: metrics}
We assess our CoLiGen model and compare it to other established methods in terms of realness and faithfulness, considering both image-based and point cloud representations of the synthesised data. Additionally, we evaluate through a perception model. The following paragraphs delve into each category of the metrics. \par
Evaluating the realness of synthesised data typically involves assessing the closeness between the distributions of real and synthesised data. To construct the sets of real and synthesised samples, we randomly selected 5k samples from the test sets of both real and synthesised datasets. We adopt the metrics used in unconditional Lidar point cloud synthesis \cite{10.1007/978-3-031-20050-2_2,Caccia2018DeepGM} and calculate Fréchet Inception Distance (FID) \cite{Heusel2017} and Sliced Wasserstein Distance (SWD) \cite{Karras2017ProgressiveGO} for the image-based representation, along with Jensen Shannon Distance (JSD) and Minimum Matching Distance (MMD) for the point cloud representation. 
\par
To assess the faithfulness of the simulated Lidar data, we employ Root Mean Square Error (RMSE) to quantify the distance between 2D input-output depth images. To obtain low RMSE values, input-output matching points must possess extremely close azimuth and elevation angles, resulting in their projection into the same cell. Consequently, we also compute the Chamfer Distance (CD) within the 3D domain, which is a more relaxed metric as it only relies on Euclidean distances in 3D coordinates for matching.
\par
To evaluate from the perspective of a downstream module, we measure the performance of the Rangenet++ \cite{8967762} model for the semantic segmentation task. We feed the pre-trained model with synthesised data and calculate the Intersection Over Union (IOU) and pixel-wise accuracy (PixelAcc). The Rangenet++ model's performance inherently indicates both the realism and faithfulness of the synthesised data.

\subsubsection{Implementation details}
We follow the setting proposed by Park et al. \cite{park2020cut} for contrastive learning and network architecture, with a few modifications. Specifically, we use the ResNet architecture \cite{Johnson2016}, PatchGAN \cite{isola2017image} and LSGAN \cite{Mao2016LeastSG} as $\G$, $\D$ and adversarial loss, respectively. Moreover, we select features from five evenly distributed layers of $\G$ as $\Genc$ and post-process them using a two-layer MLP with 256 units ($\Hnet$ network).
We set $\lambda_{idt}=2$, $\lambda_{nce}=1$ in Equation \ref{eq: total_loss} and $\alpha=1$ in Equation \ref{eq: gumble-sigmoid} due to the better faithfulness-realness trade-off and loss convergence.  We use the Adam optimiser with a learning rate of $5e^{-5}$ and implement linear learning rate decay of 0.5 at every 10th epoch. We set the batch size to 12 and observed training convergence at approximately 80 epochs. We implemented the networks using Pytorch Library and conducted all the experiments on NVIDIA Geforce RTX 3090 GPUs. We did not employ any data augmentation techniques, contrary to other image-to-image translation methods, as we did not observe any noticeable improvements in our case.

\begin{table*}[t]
\centering
\caption{Results of Ablation Study. We incrementally incorporate Raydrop Synthesis (RS), Contrastive Learning (CL), and Auxiliary Images (AI) into our Baseline and examine the effects. The best results are bolded. The lower the value in all metrics, the better the performance.}
\label{tab:ablation}
\begin{tblr}{
  colsep = 2.3mm,
  column{3-8} = {c},
  cell{2}{1} = {r=4}{},
  cell{6}{1} = {r=4}{},
  hline{1-2} = {-}{},
  hline{6,10} = {1-8}{},
  hline{1,10} = {-}{1pt}
}
Dataset        & Method             & FID       & SWD & JSD & MMD $\times 10$ & CD $\times 10$ & RMSE        \\
Semantic-KITTI & Baseline           & 2939                   & 1.76            & 0.434             & 0.52                      & 0.063                     & 0.508                   \\
               & Baseline + RS      & 2052     & 0.54            & 0.103     & 0.08             & 0.037   & 0.378              \\
               & Baseline + RS + CL      & \textbf{\textbf{1846}} & 0.55            & 0.076            & 0.04                      & 0.035                     & \textbf{\textbf{0.368}} \\
               & Baseline + RS + CL + AI (CoLiGen)& 1851                   & \textbf{0.45}   & \textbf{0.071}   & \textbf{0.03}             & \textbf{0.032}            & 0.376                   \\
SemanticPOSS   & Baseline          & 4046                   & 1.82            & 0.401           & 0.48                      & 0.272                     & 0.862                   \\
               & Baseline + RS & 3775             & 1.06     & 0.369             & 0.45             & 0.066   & 0.502             \\
               & Baseline + RS + CL      &  \textbf{3282}          & 0.79            & 0.240            & 0.10                      & 0.089                     & 0.238                  \\
               & Baseline + RS + CL + AI (CoLiGen) & 3365                   & \textbf{0.59}   & \textbf{0.228}   & \textbf{0.09}             & \textbf{0.063}            & \textbf{0.217}          
\end{tblr}
\end{table*}

\begin{table*}
\centering
\caption{Quantitative evaluation from the perspective of a downstream perception model, RangeNet++ \cite{8967762}. We report the per-class IOU and the overall Pixel-wise Accuracy (pixelAcc) for the simulated semantic-CARLA, translated data by our CoLiGen model, and the real semantic-KITTI dataset, as evaluated by the pre-trained RangeNet++. The best results are bolded.}
\label{tab:quant-sem-results}
\begin{tblr}{
  colsep=2mm,
  column{2-10}={c},
  hline{1-2,5} = {-}{},
  hline{1,5} = {-}{1pt},
  hline{4} = {-}{dashed}
}
Model & Car & Road & Sidewalk & Building & Fence & Vegetation & Terrain & Pole & Traffic-sign & PixelAcc \\
Semantic-CARLA (simulated) & 0.45 & 0.46 & \textbf{0.094} & 0.077 & 0.12 & \textbf{0.31} & 0.10 & \textbf{0.21} & 0.0007 & 0.43 \\
Translated by our CoLiGen & \textbf{0.64} & \textbf{0.70} & 0.068 & \textbf{0.125} & \textbf{0.17} & 0.21 & \textbf{0.32} & 0.19 & \textbf{0.0079} & \textbf{0.61} \\
Semantic-KITTI (real) & 0.73 & 0.85 & 0.653 & 0.614 & 0.046 & 0.66 & 0.60 & 0.16 & 0.0766 & 0.81 \\
\end{tblr}
\end{table*}

\subsection{Comparison to State-of-the-Art Models} \label{subsec:comparison to SOTA}
In this section, we assess the performance of our CoLiGen using the realness and faithfulness metrics discussed in Section \ref{sec: metrics} and compare it to SOTA controllable Lidar point cloud generative models. Since all hybrid methods \cite{9339933,Wu2018,guillard2022learning,Manivasagam2020} explained in Section \ref{related-work} employ the same technique of adapting the U-Net model \cite{Ronneberger2015} to predict reflectance or raydrop, we select one model from this category for comparison, namely RINet (Raydrop and Intensity Network) \cite{guillard2022learning}. For end-to-end models, some approaches \cite{Saleh2019DomainAF, 10.1109/ITSC48978.2021.9564553} use CycleGAN \cite{cyclegan} for domain mapping on BEV representations. As a representative of these models, we trained the vanilla CycleGAN \cite{cyclegan} model on the image-based representation. We refer to this category of models as CycleGAN. Additionally, we compare our model to the SOTA image-to-image translation model, GcGAN \cite{8953926}. We adapt both CycleGAN and GcGAN to synthesise reflectance and raydrop for a fair comparison. We also measure the realness metrics for the simulated (referred to as Semantic-CARLA) and real training (referred to as Training Set) datasets themselves which act as a lower bound and upper bound of the metrics, respectively.

\par
As reported in Table \ref{tab:sota}, our CoLiGen attains the top performance in nearly all the metrics. It is evident that our CoLiGen has successfully enhanced the realism of the simulated data, semantic-CARLA, by up to 80\% in metrics such as SWD, thereby significantly narrowing the gap to the real data domain. CycleGAN and GcGAN achieve a comparable result to ours in terms of realness in image-based representation, e.g. SWD and FID. This can be attributed to the presence of GAN loss, which pushes the images to have a real appearance.  On the other hand, RINet demonstrates comparable, or in some cases superior, performance in terms of realness in point cloud representation and overall faithfulness. We speculate that this may be attributed to the RINet's exclusive use of the L1 loss, which prevents modifications in 3D shape. \par
We visualise the synthesised depth images and the reconstructed point clouds of each method in Fig. \ref{fig:qual_comp}. Additionally, we plot the simulated input data (labelled `Simulated') and the nearest neighbour real data corresponding to the synthesised depth images (labelled `NN-Real'). From the depth images, it is evident that the CycleGAN and GcGAN models exhibit hallucination artefacts, such as adding or removing objects from the scene. Moreover, RINet fails to produce a realistic shape, which is more apparent in the point cloud representation, due to the lack of adversarial training. In contrast, our CoLiGen not only synthesises realistic Lidar point clouds but also faithfully represents the simulated input.

\subsection{Ablation Study}

We conduct an ablation study to investigate the impact of the proposed techniques in CoLiGen on the performance. As a baseline model, we select the vanilla CycleGAN network \cite{cyclegan} without the RS module. Our study encompasses three steps: firstly, the integration of RS into the baseline; secondly, replacing cycle-consistency with Contrastive Learning (CL), i.e. PatchNCE loss; and thirdly, the incorporation of Auxiliary Images (AI) into the input. As shown in Table \ref{tab:ablation}, adding RS to the baseline leads to a substantial increase in performance. This is because CNNs inherently filter high frequency in the images; Hence, disentangling the synthesis of the raydrop mask from Lidar scans can improve all the metrics significantly. Including CL in the second step significantly improves the realness metrics such as FID, JSD, and MMD, while most other metrics remain roughly unchanged. This improvement can be attributed to a more stringent consistency objective, PatchNCE, leading to a better solution.  Regarding the third step, the incorporation of AI shows to have a 5-10\% improvement in most of the metrics. It is worth mentioning that we did not observe significant improvement while using RGB images; therefore, the auxiliary images in all experiments only contained the semantic segmentation layout.

\subsection{Evaluation by a Downstream Perception Model} \label{subsec:impact on downstrean}

Since the Lidar point clouds are mainly processed by downstream perception models, we evaluate CoLiGen's synthesis quality from the perspective of a semantic segmentation model. In particular,  we predict the semantic labels of our translated point clouds using the pre-trained Rangenet++ \cite{8967762} (on the semantic-KITTI training set) and report the performance using the per-class IOU and PixelAcc. We follow the same procedure for the simulated semantic-CARLA and the real dataset, i.e. semantic-KITTI test set, with the results presented in different rows of Table \ref{tab:quant-sem-results}. As shown in the table,  our CoLiGen improves the IOU and PixelAcc for most classes, implying that our model generates more realistic objects from the perspective of the Rangenet++ model. However, performance degradation is observed in some classes, which can be attributed to the significant differences in proportions and regional distribution between the semantic-CARLA and semantic-KITTI datasets for those classes, such as Vegetation. This discrepancy causes the pre-trained RangeNet++ to make biased predictions over certain regions, leading to large errors in the segmentation. The translation of class regional distribution while preserving faithfulness continues to be an open problem and a promising avenue for future research in this field, as we will discuss in Section \ref{sec:conclusion}.\par
\begin{figure}[t!]
\centering
\includegraphics[width=\linewidth]{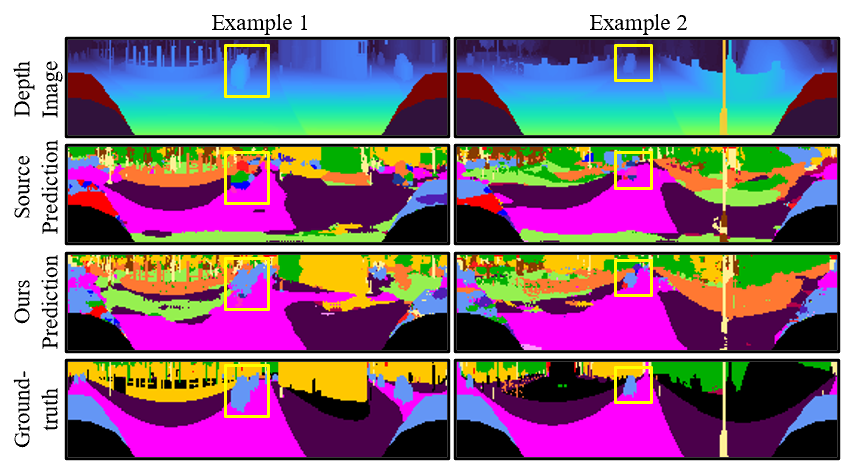}
\caption{Qualitative evaluation from the perspective of a downstream perception model, RangeNet++ \cite{8967762}. From top to bottom, the rows display the depth image of the scene, the predicted layout for the source prediction (on the simulated point cloud), the predicted layout for our translated point cloud, and the ground-truth layout. }
\label{fig:sem-results}
\end{figure}

We further visualise the predicted semantic layouts for both the simulated source data and our model, comparing them to the ground truth in Fig. \ref{fig:sem-results}. Specific regions of the images, primarily containing car objects, are highlighted to better visibility of the differences. As evident, the source layout has missed the car prediction, which is substantially improved in our prediction. This indicates that our CoLiGen model has synthesised a more realistic point cloud shape, leading to more accurate predictions by the RangeNet++ model.

\section{Conclusion and Future Work} \label{sec:conclusion}
In this paper, we introduced CoLiGen, a novel controllable Lidar point cloud generative model designed to improve the fidelity of Lidar simulation.  As a part of this research, we also created a synthetic Lidar dataset, namely the Semantic-CARLA dataset. The performance of CoLiGen was evaluated using a variety of metrics that assess realness, faithfulness, and the performance of the downstream perception task. The evaluations emphasised the significant impact of contrastive learning and the raydrop synthesis module on enhancing overall performance. Furthermore, the results demonstrated that CoLiGen not only generates realistic data but also faithfully represents the simulation, surpassing SOTA models in nearly all metrics.
\par
Our research has also revealed several critical challenges for future studies, particularly in minimising bias in regional class distributions during translation while maintaining the output's realism according to the discriminator. Achieving this balance is a crucial area for future investigation. We suggest that employing patch-wise translation methods in conjunction with sampling matching techniques \cite{9756256} could potentially address these challenges, though validating these approaches is beyond the scope of the current study. We expect this work and future research to inspire studies that bridge the gap between simulated and real-world Lidar data, improving Lidar simulation fidelity and utility in autonomous driving systems.
\bibliographystyle{IEEEtran}
\bibliography{references.bib}

\begin{thebibliography}{10}
\providecommand{\url}[1]{#1}
\csname url@samestyle\endcsname
\providecommand{\newblock}{\relax}
\providecommand{\bibinfo}[2]{#2}
\providecommand{\BIBentrySTDinterwordspacing}{\spaceskip=0pt\relax}
\providecommand{\BIBentryALTinterwordstretchfactor}{4}
\providecommand{\BIBentryALTinterwordspacing}{\spaceskip=\fontdimen2\font plus
\BIBentryALTinterwordstretchfactor\fontdimen3\font minus \fontdimen4\font\relax}
\providecommand{\BIBforeignlanguage}[2]{{%
\expandafter\ifx\csname l@#1\endcsname\relax
\typeout{** WARNING: IEEEtran.bst: No hyphenation pattern has been}%
\typeout{** loaded for the language `#1'. Using the pattern for}%
\typeout{** the default language instead.}%
\else
\language=\csname l@#1\endcsname
\fi
#2}}
\providecommand{\BIBdecl}{\relax}
\BIBdecl

\bibitem{Kalra2016}
N.~Kalra and S.~Paddock, \emph{Driving to Safety: How Many Miles of Driving Would It Take to Demonstrate Autonomous Vehicle Reliability?}\hskip 1em plus 0.5em minus 0.4em\relax RAND Corporation, 5 2016.

\bibitem{shift}
T.~Sun, M.~Segu, J.~Postels, Y.~Wang, L.~Van~Gool, B.~Schiele, F.~Tombari, and F.~Yu, ``{SHIFT:} a synthetic driving dataset for continuous multi-task domain adaptation,'' in \emph{Computer Vision and Pattern Recognition}, 2022.

\bibitem{s19030648}
\BIBentryALTinterwordspacing
F.~Rosique, P.~J. Navarro, C.~Fernández, and A.~Padilla, ``A systematic review of perception system and simulators for autonomous vehicles research,'' \emph{Sensors}, vol.~19, no.~3, 2019. [Online]. Available: \url{https://www.mdpi.com/1424-8220/19/3/648}
\BIBentrySTDinterwordspacing

\bibitem{Kadian2019Sim2RealPD}
\BIBentryALTinterwordspacing
A.~Kadian, J.~Truong, A.~Gokaslan, A.~Clegg, E.~Wijmans, S.~Lee, M.~Savva, S.~Chernova, and D.~Batra, ``Sim2real predictivity: Does evaluation in simulation predict real-world performance?'' \emph{IEEE Robotics and Automation Letters}, vol.~5, pp. 6670--6677, 2019. [Online]. Available: \url{https://api.semanticscholar.org/CorpusID:221082834}
\BIBentrySTDinterwordspacing

\bibitem{10.1002/aaai.12141}
\BIBentryALTinterwordspacing
A.~Baheri, ``Exploring the role of simulator fidelity in the safety validation of learning‐enabled autonomous systems,'' \emph{AI Mag.}, vol.~44, no.~4, p. 453–459, dec 2023. [Online]. Available: \url{https://doi.org/10.1002/aaai.12141}
\BIBentrySTDinterwordspacing

\bibitem{Triess2021ASO}
\BIBentryALTinterwordspacing
L.~T. Triess, M.~Dreissig, C.~B. Rist, and J.~M. Z{\"o}llner, ``A survey on deep domain adaptation for lidar perception,'' \emph{2021 IEEE Intelligent Vehicles Symposium Workshops (IV Workshops)}, pp. 350--357, 2021. [Online]. Available: \url{https://api.semanticscholar.org/CorpusID:235352671}
\BIBentrySTDinterwordspacing

\bibitem{9564034}
A.~Elmquist and D.~Negrut, ``Modeling cameras for autonomous vehicle and robot simulation: An overview,'' \emph{IEEE Sensors Journal}, vol.~21, no.~22, pp. 25\,547--25\,560, 2021.

\bibitem{9339933}
P.~Vacek, O.~Jašek, K.~Zimmermann, and T.~Svoboda, ``Learning to predict lidar intensities,'' \emph{IEEE Transactions on Intelligent Transportation Systems}, vol.~23, no.~4, pp. 3556--3564, 2022.

\bibitem{Ronneberger2015}
O.~Ronneberger, P.~Fischer, and T.~Brox, ``{U-net: Convolutional networks for biomedical image segmentation},'' in \emph{Lecture Notes in Computer Science (including subseries Lecture Notes in Artificial Intelligence and Lecture Notes in Bioinformatics)}, vol. 9351.\hskip 1em plus 0.5em minus 0.4em\relax Springer Verlag, 2015, pp. 234--241.

\bibitem{Geiger2013}
A.~Geiger, P.~Lenz, C.~Stiller, and R.~Urtasun, ``{Vision meets robotics: The KITTI dataset},'' \emph{International Journal of Robotics Research}, vol.~32, no.~11, pp. 1231--1237, sep 2013.

\bibitem{Dosovitskiy2017}
A.~Dosovitskiy, G.~Ros, F.~Codevilla, A.~L{\'{o}}pez, and V.~Koltun, ``{CARLA: An Open Urban Driving Simulator},'' Tech. Rep., oct 2017.

\bibitem{Saleh2019DomainAF}
\BIBentryALTinterwordspacing
K.~Saleh, A.~Abobakr, M.~Attia, J.~Iskander, D.~Nahavandi, and M.~Hossny, ``Domain adaptation for vehicle detection from bird's eye view lidar point cloud data,'' \emph{2019 IEEE/CVF International Conference on Computer Vision Workshop (ICCVW)}, pp. 3235--3242, 2019. [Online]. Available: \url{https://api.semanticscholar.org/CorpusID:162168643}
\BIBentrySTDinterwordspacing

\bibitem{10.1109/ITSC48978.2021.9564553}
\BIBentryALTinterwordspacing
A.~Barrera, J.~Beltr\'{a}n, C.~Guindel, J.~A. Iglesias, and F.~Garc\'{\i}a, ``Cycle and semantic consistent adversarial domain adaptation for reducing simulation-to-real domain shift in lidar bird's eye view,'' in \emph{2021 IEEE International Intelligent Transportation Systems Conference (ITSC)}.\hskip 1em plus 0.5em minus 0.4em\relax IEEE Press, 2021, p. 3081–3086. [Online]. Available: \url{https://doi.org/10.1109/ITSC48978.2021.9564553}
\BIBentrySTDinterwordspacing

\bibitem{cyclegan}
J.-Y. Zhu, T.~Park, P.~Isola, and A.~A. Efros, ``Unpaired image-to-image translation using cycle-consistent adversarial networks,'' in \emph{2017 IEEE International Conference on Computer Vision (ICCV)}, 2017, pp. 2242--2251.

\bibitem{xiao2022transfer}
A.~Xiao, J.~Huang, D.~Guan, F.~Zhan, and S.~Lu, ``Transfer learning from synthetic to real lidar point cloud for semantic segmentation,'' in \emph{Proceedings of the AAAI Conference on Artificial Intelligence}, vol.~36, no.~3, 2022, pp. 2795--2803.

\bibitem{Nakashima2021LearningTD}
\BIBentryALTinterwordspacing
K.~Nakashima and R.~Kurazume, ``Learning to drop points for lidar scan synthesis,'' \emph{2021 IEEE/RSJ International Conference on Intelligent Robots and Systems (IROS)}, pp. 222--229, 2021. [Online]. Available: \url{https://api.semanticscholar.org/CorpusID:232035774}
\BIBentrySTDinterwordspacing

\bibitem{park2020cut}
T.~Park, A.~A. Efros, R.~Zhang, and J.-Y. Zhu, ``Contrastive learning for unpaired image-to-image translation,'' in \emph{European Conference on Computer Vision}, 2020.

\bibitem{Heusel2017}
M.~Heusel, H.~Ramsauer, T.~Unterthiner, B.~Nessler, and S.~Hochreiter, ``{GANs Trained by a Two Time-Scale Update Rule Converge to a Local Nash Equilibrium},'' \emph{Advances in Neural Information Processing Systems}, vol. 2017-December, pp. 6627--6638, jun 2017.

\bibitem{10.5555/2969033.2969125}
I.~J. Goodfellow, J.~Pouget-Abadie, M.~Mirza, B.~Xu, D.~Warde-Farley, S.~Ozair, A.~Courville, and Y.~Bengio, ``Generative adversarial nets,'' in \emph{Proceedings of the 27th International Conference on Neural Information Processing Systems - Volume 2}, ser. NIPS'14.\hskip 1em plus 0.5em minus 0.4em\relax Cambridge, MA, USA: MIT Press, 2014, p. 2672–2680.

\bibitem{10.5555/3495724.3496298}
J.~Ho, A.~Jain, and P.~Abbeel, ``Denoising diffusion probabilistic models,'' in \emph{Proceedings of the 34th International Conference on Neural Information Processing Systems}, ser. NIPS '20.\hskip 1em plus 0.5em minus 0.4em\relax Red Hook, NY, USA: Curran Associates Inc., 2020.

\bibitem{10.5555/3295222.3295349}
A.~Vaswani, N.~Shazeer, N.~Parmar, J.~Uszkoreit, L.~Jones, A.~N. Gomez, L.~Kaiser, and I.~Polosukhin, ``Attention is all you need,'' in \emph{Proceedings of the 31st International Conference on Neural Information Processing Systems}, ser. NIPS'17.\hskip 1em plus 0.5em minus 0.4em\relax Red Hook, NY, USA: Curran Associates Inc., 2017, p. 6000–6010.

\bibitem{Caccia2018DeepGM}
\BIBentryALTinterwordspacing
L.~Caccia, H.~van Hoof, A.~C. Courville, and J.~Pineau, ``Deep generative modeling of lidar data,'' \emph{2019 IEEE/RSJ International Conference on Intelligent Robots and Systems (IROS)}, pp. 5034--5040, 2018. [Online]. Available: \url{https://api.semanticscholar.org/CorpusID:54445260}
\BIBentrySTDinterwordspacing

\bibitem{Radford2015UnsupervisedRL}
\BIBentryALTinterwordspacing
A.~Radford, L.~Metz, and S.~Chintala, ``Unsupervised representation learning with deep convolutional generative adversarial networks,'' \emph{CoRR}, vol. abs/1511.06434, 2015. [Online]. Available: \url{https://api.semanticscholar.org/CorpusID:11758569}
\BIBentrySTDinterwordspacing

\bibitem{nakashima2022generative}
K.~Nakashima, Y.~Iwashita, and R.~Kurazume, ``Generative range imaging for learning scene priors of 3d lidar data,'' 2022.

\bibitem{10.1007/978-3-031-20050-2_2}
\BIBentryALTinterwordspacing
V.~Zyrianov, X.~Zhu, and S.~Wang, ``Generate realistic lidar point clouds,'' in \emph{Computer Vision – ECCV 2022: 17th European Conference, Tel Aviv, Israel, October 23–27, 2022, Proceedings, Part XXIII}.\hskip 1em plus 0.5em minus 0.4em\relax Berlin, Heidelberg: Springer-Verlag, 2022, p. 17–35. [Online]. Available: \url{https://doi.org/10.1007/978-3-031-20050-2_2}
\BIBentrySTDinterwordspacing

\bibitem{nakashima2024lidar}
K.~Nakashima and R.~Kurazume, ``Lidar data synthesis with denoising diffusion probabilistic models,'' in \emph{Proceedings of the International Conference on Robotics and Automation (ICRA)}, 2024, pp. 14\,724--14\,731.

\bibitem{hu2024rangeldm}
Q.~Hu, Z.~Zhang, and W.~Hu, ``Rangeldm: Fast realistic lidar point cloud generation,'' 2024.

\bibitem{ran2024towards}
H.~Ran, V.~Guizilini, and Y.~Wang, ``Towards realistic scene generation with lidar diffusion models,'' in \emph{Proceedings of the IEEE/CVF Conference on Computer Vision and Pattern Recognition}, 2024.

\bibitem{NIPS2017_7a98af17}
\BIBentryALTinterwordspacing
A.~van~den Oord, O.~Vinyals, and k.~kavukcuoglu, ``Neural discrete representation learning,'' in \emph{Advances in Neural Information Processing Systems}, I.~Guyon, U.~V. Luxburg, S.~Bengio, H.~Wallach, R.~Fergus, S.~Vishwanathan, and R.~Garnett, Eds., vol.~30.\hskip 1em plus 0.5em minus 0.4em\relax Curran Associates, Inc., 2017. [Online]. Available: \url{https://proceedings.neurips.cc/paper_files/paper/2017/file/7a98af17e63a0ac09ce2e96d03992fbc-Paper.pdf}
\BIBentrySTDinterwordspacing

\bibitem{xiong2023learning}
Y.~Xiong, W.-C. Ma, J.~Wang, and R.~Urtasun, ``Learning compact representations for lidar completion and generation,'' in \emph{CVPR}, 2023.

\bibitem{haghighi2024taming}
H.~Haghighi, A.~Samadi, M.~Dianati, V.~Donzella, and K.~Debattista, ``Taming transformers for realistic lidar point cloud generation,'' 2024.

\bibitem{Wu2018}
B.~Wu, X.~Zhou, S.~Zhao, X.~Yue, and K.~Keutzer, ``Squeezesegv2: Improved model structure and unsupervised domain adaptation for road-object segmentation from a lidar point cloud,'' \emph{Proceedings - IEEE International Conference on Robotics and Automation}, vol. 2019-May, pp. 4376--4382, 9 2018.

\bibitem{guillard2022learning}
B.~Guillard, S.~Vemprala, J.~K. Gupta, O.~Miksik, V.~Vineet, P.~Fua, and A.~Kapoor, ``Learning to simulate realistic lidars,'' 2022.

\bibitem{Zhao2020ePointDAAE}
\BIBentryALTinterwordspacing
S.~Zhao, Y.~Wang, B.~Li, B.~Wu, Y.~Gao, P.~Xu, T.~Darrell, and K.~Keutzer, ``epointda: An end-to-end simulation-to-real domain adaptation framework for lidar point cloud segmentation,'' in \emph{AAAI Conference on Artificial Intelligence}, 2020. [Online]. Available: \url{https://api.semanticscholar.org/CorpusID:221534728}
\BIBentrySTDinterwordspacing

\bibitem{Manivasagam2020}
S.~Manivasagam, S.~Wang, K.~Wong, W.~Zeng, M.~Sazanovich, S.~Tan, B.~Yang, W.~C. Ma, and R.~Urtasun, ``{LiDARsim: Realistic LiDAR simulation by leveraging the real world},'' in \emph{Proceedings of the IEEE Computer Society Conference on Computer Vision and Pattern Recognition}.\hskip 1em plus 0.5em minus 0.4em\relax IEEE Computer Society, jun 2020, pp. 11\,164--11\,173.

\bibitem{Sallab2019}
A.~E. Sallab, I.~Sobh, M.~Zahran, and M.~Shawky, ``{Unsupervised Neural Sensor Models for Synthetic LiDAR Data Augmentation},'' \emph{arXiv}, nov 2019.

\bibitem{nakashima2021learning}
K.~Nakashima and R.~Kurazume, ``Learning to drop points for lidar scan synthesis,'' \emph{arXiv preprint arXiv:2102.11952}, 2021.

\bibitem{8953926}
H.~Fu, M.~Gong, C.~Wang, K.~Batmanghelich, K.~Zhang, and D.~Tao, ``Geometry-consistent generative adversarial networks for one-sided unsupervised domain mapping,'' in \emph{2019 IEEE/CVF Conference on Computer Vision and Pattern Recognition (CVPR)}, 2019, pp. 2422--2431.

\bibitem{behley2019iccv}
J.~Behley, M.~Garbade, A.~Milioto, J.~Quenzel, S.~Behnke, C.~Stachniss, and J.~Gall, ``{SemanticKITTI: A Dataset for Semantic Scene Understanding of LiDAR Sequences},'' in \emph{Proc. of the IEEE/CVF International Conf.~on Computer Vision (ICCV)}, 2019.

\bibitem{pan2020semanticposs}
Y.~Pan, B.~Gao, J.~Mei, S.~Geng, C.~Li, and H.~Zhao, ``Semanticposs: A point cloud dataset with large quantity of dynamic instances,'' 2020.

\bibitem{nuscenes}
H.~Caesar, V.~Bankiti, A.~H. Lang, S.~Vora, V.~E. Liong, Q.~Xu, A.~Krishnan, Y.~Pan, G.~Baldan, and O.~Beijbom, ``nuscenes: A multimodal dataset for autonomous driving,'' in \emph{CVPR}, 2020.

\bibitem{Sun_2020_CVPR}
P.~Sun, H.~Kretzschmar, X.~Dotiwalla, A.~Chouard, V.~Patnaik, P.~Tsui, J.~Guo, Y.~Zhou, Y.~Chai, B.~Caine, V.~Vasudevan, W.~Han, J.~Ngiam, H.~Zhao, A.~Timofeev, S.~Ettinger, M.~Krivokon, A.~Gao, A.~Joshi, Y.~Zhang, J.~Shlens, Z.~Chen, and D.~Anguelov, ``Scalability in perception for autonomous driving: Waymo open dataset,'' in \emph{Proceedings of the IEEE/CVF Conference on Computer Vision and Pattern Recognition (CVPR)}, June 2020.

\bibitem{song2023synthetic}
Z.~Song, Z.~He, X.~Li, Q.~Ma, R.~Ming, Z.~Mao, H.~Pei, L.~Peng, J.~Hu, D.~Yao, and Y.~Zhang, ``Synthetic datasets for autonomous driving: A survey,'' 2023.

\bibitem{10.1109/IVS.2019.8813809}
\BIBentryALTinterwordspacing
B.~Hurl, K.~Czarnecki, and S.~Waslander, ``Precise synthetic image and lidar (presil) dataset for autonomous vehicle perception,'' in \emph{2019 IEEE Intelligent Vehicles Symposium (IV)}.\hskip 1em plus 0.5em minus 0.4em\relax IEEE Press, 2019, p. 2522–2529. [Online]. Available: \url{https://doi.org/10.1109/IVS.2019.8813809}
\BIBentrySTDinterwordspacing

\bibitem{Weng2020_AIODrive}
X.~Weng, Y.~Man, J.~Park, Y.~Yuan, D.~Cheng, M.~O'Toole, and K.~Kitani, ``{All-In-One Drive: A Large-Scale Comprehensive Perception Dataset with High-Density Long-Range Point Clouds},'' \emph{arXiv}, 2021.

\bibitem{alotaibi2020deep}
A.~Alotaibi, ``Deep generative adversarial networks for image-to-image translation: A review,'' \emph{Symmetry}, vol.~12, no.~10, p. 1705, 2020.

\bibitem{Bora2018AmbientGANGM}
\BIBentryALTinterwordspacing
A.~Bora, E.~Price, and A.~G. Dimakis, ``Ambientgan: Generative models from lossy measurements,'' in \emph{International Conference on Learning Representations}, 2018. [Online]. Available: \url{https://api.semanticscholar.org/CorpusID:3481010}
\BIBentrySTDinterwordspacing

\bibitem{journals/corr/JangGP16}
\BIBentryALTinterwordspacing
E.~Jang, S.~Gu, and B.~Poole, ``Categorical reparameterization with gumbel-softmax.'' \emph{CoRR}, vol. abs/1611.01144, 2016. [Online]. Available: \url{http://dblp.uni-trier.de/db/journals/corr/corr1611.html#JangGP16}
\BIBentrySTDinterwordspacing

\bibitem{Karras2017ProgressiveGO}
\BIBentryALTinterwordspacing
T.~Karras, T.~Aila, S.~Laine, and J.~Lehtinen, ``Progressive growing of gans for improved quality, stability, and variation,'' \emph{ArXiv}, vol. abs/1710.10196, 2017. [Online]. Available: \url{https://api.semanticscholar.org/CorpusID:3568073}
\BIBentrySTDinterwordspacing

\bibitem{8967762}
A.~Milioto, I.~Vizzo, J.~Behley, and C.~Stachniss, ``Rangenet ++: Fast and accurate lidar semantic segmentation,'' in \emph{2019 IEEE/RSJ International Conference on Intelligent Robots and Systems (IROS)}, 2019, pp. 4213--4220.

\bibitem{Johnson2016}
J.~Johnson, A.~Alahi, and L.~Fei-Fei, ``{Perceptual Losses for Real-Time Style Transfer and Super-Resolution},'' \emph{Lecture Notes in Computer Science (including subseries Lecture Notes in Artificial Intelligence and Lecture Notes in Bioinformatics)}, vol. 9906 LNCS, pp. 694--711, mar 2016.

\bibitem{isola2017image}
P.~Isola, J.-Y. Zhu, T.~Zhou, and A.~A. Efros, ``Image-to-image translation with conditional adversarial networks,'' in \emph{Proceedings of the IEEE conference on computer vision and pattern recognition}, 2017, pp. 1125--1134.

\bibitem{Mao2016LeastSG}
\BIBentryALTinterwordspacing
X.~Mao, Q.~Li, H.~Xie, R.~Y.~K. Lau, Z.~Wang, and S.~P. Smolley, ``Least squares generative adversarial networks,'' \emph{2017 IEEE International Conference on Computer Vision (ICCV)}, pp. 2813--2821, 2016. [Online]. Available: \url{https://api.semanticscholar.org/CorpusID:206771128}
\BIBentrySTDinterwordspacing

\bibitem{9756256}
S.~R. Richter, H.~Alhaija, and V.~Koltun, ``Enhancing photorealism enhancement,'' \emph{IEEE Transactions on Pattern Analysis \&amp; Machine Intelligence}, vol.~45, no.~02, pp. 1700--1715, feb 2023.

\end{thebibliography}
\begin{IEEEbiography}[{\includegraphics[width=1in,height=1.25in,clip,keepaspectratio]{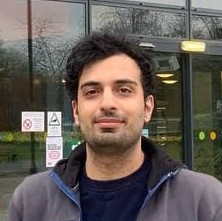}}]{Hamed Haghighi}
is a PhD candidate with the Warwick Manufacturing Group (WMG) at University
of Warwick, UK. He received a B.Sc. (2016) in Software Engineering from Isfahan University of Technology (Isfahan, Iran) and an M.Sc. (2019) in Artificial Intelligence from University of Tehran (Tehran, Iran). His research interests include machine learning, computer
vision, computer graphics, and autonomous vehicles.
\end{IEEEbiography}
\vskip -1\baselineskip plus -1fil
\begin{IEEEbiography}[{\includegraphics[width=0.9in,clip,keepaspectratio]{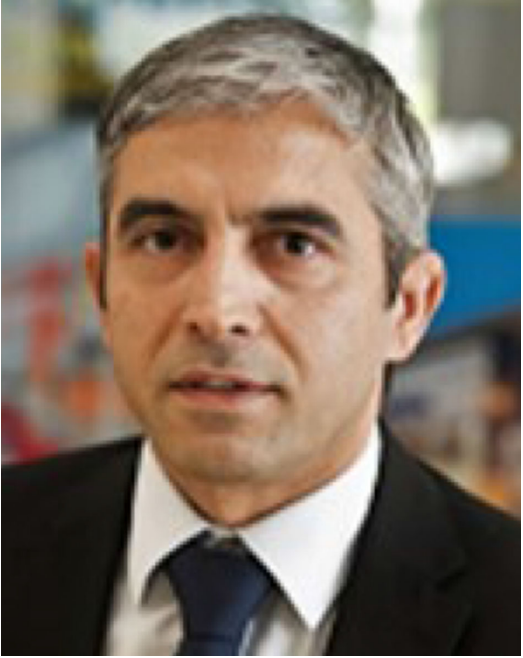}}]{Mehrdad Dianati}
(Senior Member, IEEE) is a Connected and Cooperative Autonomous Systems Professor at the School of Electrical, Electronics Engineering and Computer Science and the Director of Industry Engagement for the EPSRC Doctoral Training Centre, VDT-FORT, at Queen's University Belfast, UK. Previously, he was the Head of the Intelligent Vehicles Research Directorate and Head of Networked Intelligent Systems (Cooperative Autonomy) research at
Warwick Manufacturing Group (WMG) and the Director of Centre for Doctoral Training in Future Mobility Technologies of the University of Warwick, UK. He has over 30 years of combined industrial and academic experience, with 25 years in leadership roles in multi-disciplinary collaborative R\&D projects in close collaboration with the Automotive and ICT industries. Prof Dianati is the Field Chief Editor of Frontiers Journal in Future Transportation. Previously, he was an Associate Editor for the IEEE Transactions on Vehicular Technology and Guest Editor of several other IEEE and IET journals.
\end{IEEEbiography}
\vskip -2\baselineskip plus -1fil
\begin{IEEEbiography}[{\includegraphics[width=1in,height=1.25in,clip,keepaspectratio]{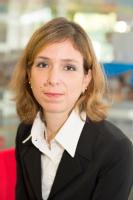}}]{Valentina Donzella}
received her BSc (2003) and MSc (2005) in
Electronics Engineering from University of Pisa and Sant’Anna School of
Advanced Studies (Pisa, Italy), and her PhD (2010) in Innovative
Technologies for Information, Communication and Perception Engineering
from Sant’Anna School of Advanced Studies. In 2009, she was a visiting
graduate student at McMaster University (Hamilton, ON, Canada) in the
Engineering Physics department.
She is currently full Professor, and Head of the Intelligent Vehicles - sensors group at
WMG, University of Warwick, UK; before this position, she was a MITACS
and SiEPIC postdoctoral fellow at the University of British Columbia
(Vancouver, BC, Canada), in the Silicon Photonics group. She is the first author
and co-author of several journal papers in top-tier optics journals. Her
research interests are LiDAR, Intelligent Vehicles, integrated optical sensors,
sensor fusion, and silicon photonics.
Dr Donzella is full College member of EPSRC and a Senior Fellow of
Higher Education Academy.
\end{IEEEbiography}
\vskip -2\baselineskip plus -1fil
\begin{IEEEbiography}[{\includegraphics[width=1in,height=1.25in,clip,keepaspectratio]{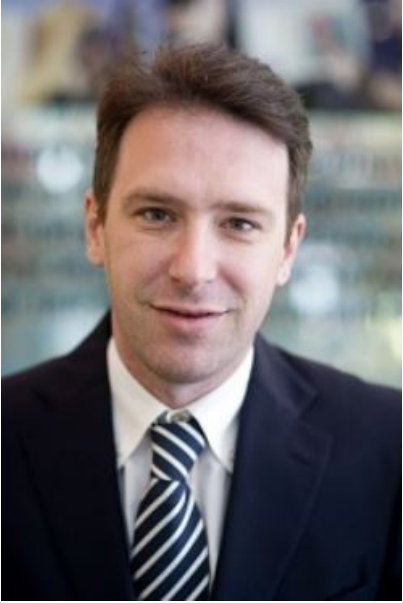}}]{Kurt Debattista}
received a B.Sc. in mathematics
and computer science, an M.Sc. in psychology, an
M.Sc. degree in computer science, and a Ph.D. in Computer Science from
the University of Bristol. He is currently a Professor
with WMG, at the University of Warwick. His
research interests are high-fidelity rendering, HDR
imaging, machine learning, and applied perception.
\end{IEEEbiography}

\end{document}